\pdfoutput=1
\documentclass[11pt]{article}

\usepackage[]{ACL2023}

\usepackage{times}
\usepackage{latexsym}

\usepackage[T1]{fontenc}
\usepackage[utf8]{inputenc}

\usepackage{microtype}

\usepackage{inconsolata}

\usepackage{graphicx}
\usepackage{multirow}
\usepackage{colortbl}
\usepackage{amsmath}
\usepackage{comment}
\usepackage{array}
\usepackage{amsfonts,amssymb}
\usepackage{tabularx}
\usepackage{booktabs}
\usepackage{subcaption}
\usepackage{float}

\newcommand{\tabincell}[2]{\begin{tabular}{@{}#1@{}}#2\end{tabular}}

\title{$\boldsymbol{M^2}$-Encoder: Advancing Bilingual Image-Text Understanding \\ by Large-scale Efficient Pretraining}

\author{
Qingpei Guo\thanks{\ \ Co-first authors} \thanks{\ \ Corresponding authors} , {\bf Furong Xu\footnotemark[1]} , {\bf Hanxiao Zhang\footnotemark[1]} , {\bf Wang Ren\footnotemark[1]} , \\
{\bf Ziping Ma}, {\bf Lin Ju}, {\bf Jian Wang}, {\bf Jingdong Chen}, {\bf Ming Yang\footnotemark[2]} 
\\ Ant Group 
\\ \{qingpei.gqp, m.yang\}@antgroup.com
} 
\begin{document}
\maketitle
\begin{abstract}
Vision-language foundation models like CLIP have revolutionized the field of artificial intelligence. Nevertheless, VLM models supporting multi-language, \emph{e.g.}, in both Chinese and English, have lagged due to the relative scarcity of large-scale pretraining datasets. Toward this end, we introduce a comprehensive bilingual (Chinese-English) dataset BM-6B with over 6 billion image-text pairs, aimed at enhancing multimodal foundation models to well understand images in both languages. To handle such a scale of dataset, we propose a novel grouped aggregation approach for image-text contrastive loss computation, which reduces the communication overhead and GPU memory demands significantly, facilitating a 60\% increase in training speed. We pretrain a series of bilingual image-text foundation models with an enhanced fine-grained understanding ability on BM-6B, the resulting models, dubbed as $M^2$-Encoders (pronounced “M-Square”), set new benchmarks in both languages for multimodal retrieval and classification tasks. Notably, Our largest $M^2$-Encoder-10B model has achieved top-1 accuracies of 88.5\% on ImageNet and 80.7\% on ImageNet-CN under a zero-shot classification setting, surpassing previously reported SoTA methods by 2.2\% and 21.1\%, respectively. The $M^2$-Encoder series represents one of the most comprehensive bilingual image-text foundation models to date, so we are making it available to the research community for further exploration and development.\footnote{Github:https://github.com/alipay/Ant-Multi-Modal-Framework/tree/main/prj/M2\_Encoder}
\end{abstract}

\section{Introduction}

Vision-language foundation models, such as CLIP \cite{clip}, are typically developed through contrastive learning by aligning image-text pairs on large-scale unsupervised or weakly supervised datasets, establishing them as fundamental components of artificial intelligence. Benefiting from their robust visual and textual representation abilities and exceptional zero-shot transferability, they are widely used in modern large-scale multimodal models, where they serve key roles in visual understanding\cite{mplugowl,blip2, gong2023multimodal,minigpt4,instructblip,llava,qwen-vl}, and cross-modal alignment and  generation\cite{dalle2,dalle1,sd}.

\begin{figure}[tp]
    \centering
    \includegraphics[width=1\linewidth]{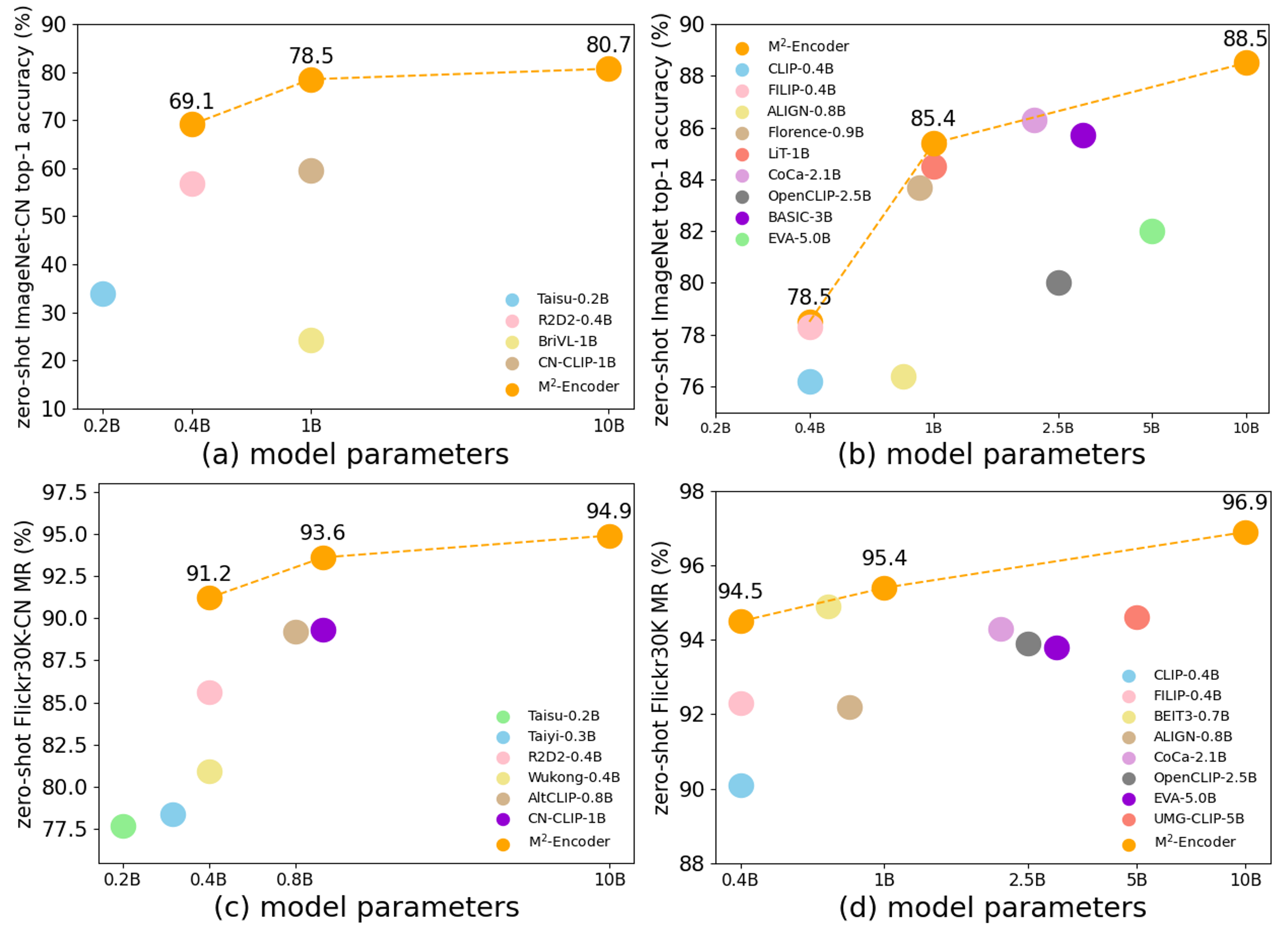}
    \caption{An overview of existing multimodal models on zero-shot classification and retrieval performance. The top-1 accuracy on (a) ImageNet-CN and (b) ImageNet. The retrieval MR on (c) Flicker30K-CN and (d) Flicker30K. Our $M^2$-Encoders excel compared to models with a similar number of parameters.}
    \label{fig:effect}
\end{figure}

The performance of image-text foundational models relies heavily on large-scale image-text datasets. However, there lack of a large-scale image-text dataset in Chinese comparable to LAION2B-EN\cite{laion5b}, which might have hindered the performance of Chinese multimodal foundational models and their real-world applications. Our work aims to narrow this gap in data scales. Toward this end, we curate image-text pairs collected from public datasets and legally sourced web content, techniques such as translation into Chinese, data cleaning to remove noise, and data augmentation to enhance variability were implemented as part of our methodology, resulting in a large-scale dataset comprising over 3 billion Chinese image-text pairs, a volume that is even larger than datasets such as LAION2B-EN. To the best of our knowledge, this collection constitutes the largest Chinese image-text dataset available to date. By integrating this corpus with English publicly available datasets(eg. LAION2B-EN, COYO-700M\cite{kakaobrain2022coyo-700m}, Datacomp-1B\cite{gadre2023datacomp}) and accounting for potential overlaps, we have constructed a high-quality bilingual dataset dubbed as BM-6B(BM represents bilingual multi-modality) that includes nearly 6 billion unique image-text pairs. The construction of this dataset provides a critical foundation for developing advanced bilingual multimodal models catering to both Chinese and English languages.

Training on such a massive dataset necessitates a substantial increase in computational resources. 
The conventional image-text contrastive (ITC) loss calculation requires gathering image-text representations from all computing nodes in a distributed system. This leads to significant communication overhead and risk of GPU memory depletion (out-of-memory errors) in large-scale training scenarios. To overcome this challenge, we design a new grouped aggregation strategy dubbed Grouped-ITC with batch accumulation(abbreviated as GBA-ITC) that evenly divides the nodes in the cluster into multiple groups. During the computation of the ITC loss, aggregation is performed within each group and coupled with batch accumulation, enabling the decoupling of the ITC loss computation from the overall batch size, leading to reduced memory requirements and enhanced scalability. This technique yields a 60\% acceleration in training speed. We also adopted the "SHARING-DELINKING" training strategy proposed by the M6-10T project\cite{m6-10t}, and utilized the ReCLIP\cite{reclip} strategy to expedite the convergence efficiency of training. 

With the aforementioned efficient training methods, we trained a series of $M^2$-Encoder models on BM-6B, with a focus on enhanced fine-grained understanding capabilities. Our $M^2$-Encoders are a series of models spanning from 0.4 billion to 10 billion parameters. We conducted zero-shot evaluations of our models' performance on six bilingual cross-modal retrieval and classification test sets, including three English test datasets: ImageNet\cite{deng2009imagenet}, Flickr30K\cite{plummer2015flickr30k}, COCO\cite{chen2015microsoft}, and three equivalent Chinese version test datasets, respectively: ImageNet-CN\cite{cnclip}, Flickr30K-CN\cite{flick30k-cn}, and COCO-CN\cite{coco-cn}. As shown in Figure \ref{fig:effect}, all of our models have achieved state-of-the-art results with comparable numbers of parameters, across multimodal retrieval and classification tasks in both Chinese and English. For fine-grained evaluation, we collect tasks requiring fine-grained perception, including fine-grained category recognition, counting, multiple object combination recognition, and relationships between objects, and established a bilingual fine-grained benchmark. Our $M^2$-Encoder-10B surpass existing CLIP-based models on our fine-grained benchmark by a large margin, with an absolute improvement of 21.58\% over CN-CLIP\cite{cnclip} for Chinese, and 15.2\% over CLIP\cite{clip} for English. Our main contributions are as follows:

\begin{itemize}
    \item We propose BM-6B, an ultra-large dataset consisting of 6 billion image-text pairs with Chinese and English data nearly equally distributed, to mitigate the shortage of extensive Chinese image-text datasets. We verify that the BM-6B dataset is large enough to facilitate the training of bilingual image-text multimodal foundational models from scratch. 
    \item We introduced a novel grouped aggregation strategy named GBA-ITC that leads to reduced memory requirements and enhanced scalability. This technique yields a 60\% acceleration in training speed, facilitating large-scale efficient pretraining.
    \item We pretrain the $M^2$-Encoder series models on the BM-6B dataset, placing additional emphasis on their fine-grained perception abilities. The resulting $M^2$-Encoder-10B model achieves SOTA performance not only across six bilingual cross-modal retrieval and classification datasets but also excels in our constructed fine-grained perception benchmark within a zero-shot learning setup.
\end{itemize}

\section{Method}
\subsection{$\boldsymbol{M^2}$-Encoder Model}
\label{sec: model}

\begin{figure*}[tph]
    \centering
    \includegraphics[width=0.8\linewidth, height=0.4\linewidth]{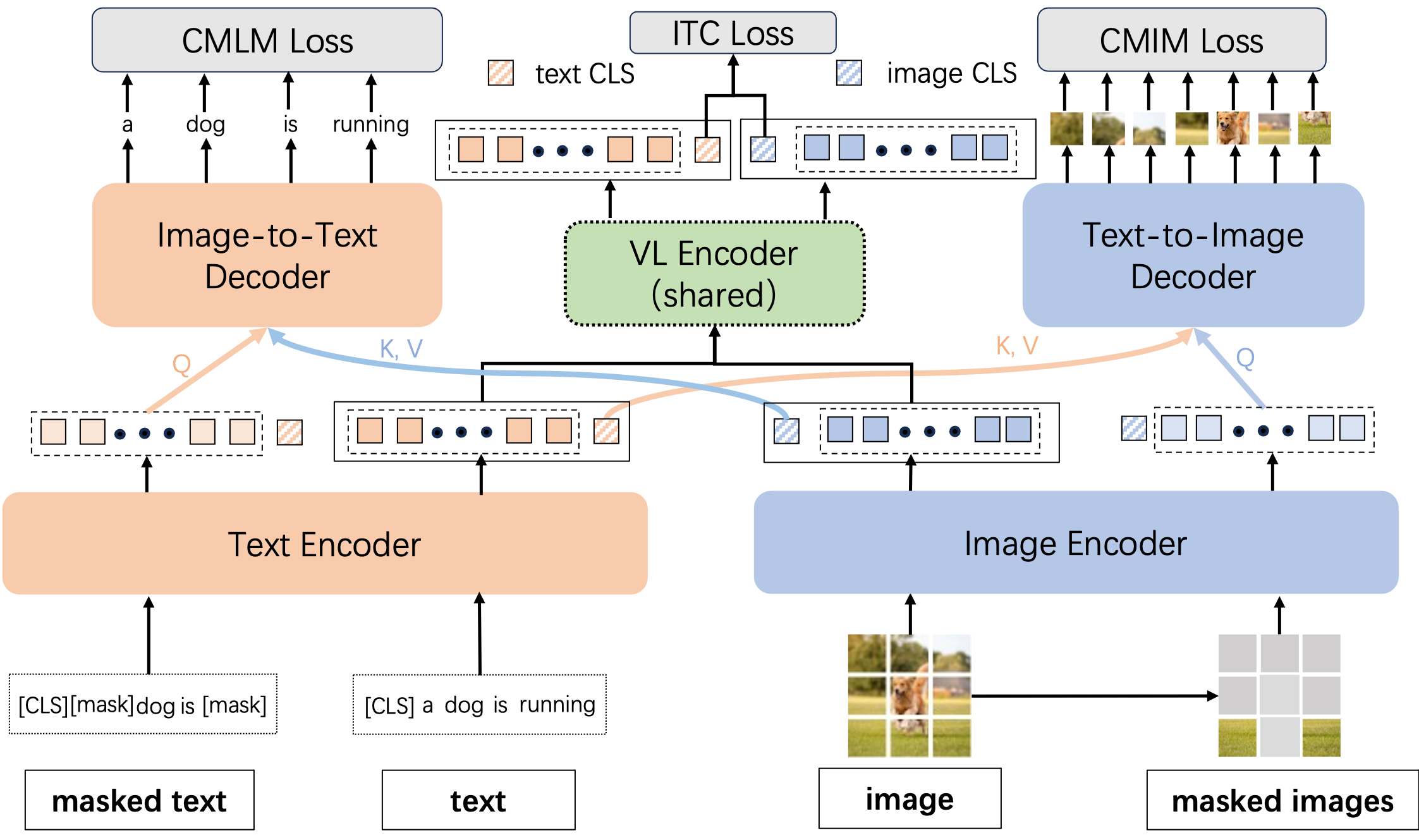}
    \caption{The pretraining tasks of the $M^2$-Encoder models, including ITC, CMLM, and CMIM heads. ITC uses [CLS] token to achieve global feature alignment between image and text pair. In contrast, the CMLM and CMIM tasks employ local tokens between modalities for cross-attention to assist masked token recovery. Together, these tasks enable the $M^2$-Encoder to align and integrate information across modalities at both a global and local level.}
    \label{fig:model_architecture}
\end{figure*}

\textbf{Model Architecture.} The availability of such a large dataset facilitates training our $M^2$-Encoder series models from scratch, allowing us to leverage cutting-edge architectures. Aiming to provide the multimodal community with highly scalable bilingual(Chinese-English) image-text foundational models, the $M^2$-Encoder series of models employ the more advanced MAGNETO\cite{magneto} transformer that is designed as a general-purpose architecture for multimodal and multitask applications. 

\textbf{Pretraining Tasks.} The pretraining tasks of $M^2$-Encoder is illustrated in Figure \ref{fig:model_architecture}. For the design of proxy tasks, we referenced SyCoCa\cite{sycoca}, which improved the CoCa pretraining tasks by introducing bidirectional interactions between images and texts. To enhance training efficiency, we opted out of the attentive masking strategy employed by SyCoCa and instead used a large proportion of random masks. Additionally, our decoder only utilized the [CLS] token from the other modality when recovering masked information.  Our $M^2$-Encoder has three heads: Image-Text Contrastive (ITC), Cross-modal Masked Language Modeling (CMLM), and Cross-modal Masked Image Masking (CMIM), with the latter two heads being exclusive to the training phase and not utilized during inference. For the ITC loss, we employ the [CLS] token from both image and text modalities for alignment. For the CMLM loss, we mask a large portion of the text and then use the image's entire token set for cross-attention via an image-to-text decoder, aiding in text token prediction. Similarly, for CMIM loss-which is analogous to CMLM—we leverage a text-to-image decoder to facilitate image recovery by inputting all text tokens and a high ratio of masked image patches. These three losses contribute to global and local cross-modal alignment, leading to improved outcomes on our constructed fine-grained benchmark.

\textbf{Training Objectives.} We present the mathematical formulations for each pretraining task. The ITC loss, designed to align text and image representations, is defined as:
\begin{align}
	\mathcal{L}_{ITC} &= \frac{1}{2N} \left[ \sum^N_{i} \log \left( \frac{\exp(\langle v^{cls}_i, t^{cls}_i\rangle/\tau)}{\sum^N_{j=1}\exp(\langle v^{cls}_i, t^{cls}_j \rangle/\tau)} \right) \right] \nonumber \\ 
	&+ \frac{1}{2N} \left[ \sum^N_{i} \log \left( \frac{\exp(\langle v^{cls}_i, t^{cls}_i\rangle/\tau)}{\sum^N_{j=1}\exp(\langle v^{cls}_i, t^{cls}_j \rangle/\tau)} \right) \right],
\end{align}
where $v^{cls}$ and $t^{cls}$ are the [CLS] tokens of image and text respectively, $N$ is the batch size. $\langle \cdot, \cdot \rangle$ denotes the inner product, and $\tau$ is a temperature parameter that scales the logits.

The CMIM loss, inspired by Masked Autoencoders (MAE)\cite{he2022masked} that reconstructs the pixels of masked image tokens, diverges from MAE by leveraging the text [CLS] token as additional context for cross-attention within the decoder. The CMIM loss function is defined as:
\begin{equation}
	\mathcal{L}_{CMIM}= \frac{1}{M} \sum_{i=0}^{P} {|| x_{i} - \hat{x}_{i} ||^2}
	\label{equ:cmim}
\end{equation}
where $M$ is the number of masked patches, $P$ is the number of pixels in patches, $x{i}$ represents the original pixel values, and $\hat{x}_{i}$ denotes the reconstructed pixel values by the decoder.

The CMLM task is modeled as a classification problem with cross-entropy, the loss is given by:
\begin{equation}
	\mathcal{L}_{CMLM} = - \frac{1}{N}\sum_{i=1}^{N} \sum_{j=1}^{Q} y_{ij} \log p_{ij}(x)
\end{equation}
where $N$ is the number of tokens in a sequence, $Q$ is the size of the vocabulary, $y_{ij}$ is the one-hot encoded true label for the i-th token, and $p_{ij}(x)$ is the predicted probability that the i-th token belongs to the j-th category.

The overall training objective for the $M^2$-Encoders combines these losses as follows:
\begin{equation}
	\mathcal{L}_{overall}=\mathcal{L}_{ITC} + \alpha \mathcal{L}_{CMIM} + \beta \mathcal{L}_{CMLM}
	\label{equ:encoder}
\end{equation}
where $\alpha$ and $\beta$ are hyperparameters that balance the relative importance of the CMIM and CMLM tasks in the overall training objective.
% TODO: alpha 和 beta 在实现细节补充

\subsection{Training Dataset}

\textbf{Data Sources.}  A significant contributor to the outstanding performance of the $M^2$-Encoder models is our construction of the large-scale BM-6B dataset. This dataset comprises approximately 6 billion bilingual image-text pairs, with each language constituting half of the dataset. The majority of these pairs were sourced from publicly available datasets. The English data within the BM-6B dataset are derived from Laion-EN\cite{laion5b}, COYO-700M\cite{kakaobrain2022coyo-700m}, Datacomp-1B\cite{gadre2023datacomp}, and Generated Datacomp's Large Pool\cite{nguyen2023improving}, while the Chinese portion is primarily sourced from Laion-CN, Wukong \cite{carver2020wukong}, Taisu \cite{liu2022taisu}, and Zero \cite{xie2023ccmb}. Additionally, to enhance the diversity of the training dataset, we incorporated 680 million Chinese in-house data samples and translated the Laion-EN dataset into Chinese. The detailed distribution of the training data is presented in Appendix \ref{app:bm_6B_distributions}.

\textbf{Data Cleaning.} Besides the quantity, the quality of data is crucial for the effectiveness of the model. To improve the quality of the BM-6B dataset, we developed a data cleaning pipeline. We first remove samples if they satisfy any of the following conditions: the text length is less than 5 characters, which typically indicates insufficient descriptive content, or the image aspect ratio exceeds 3, suggesting distorted or extreme image dimensions. For the samples that remain, we calculate the image-text semantic similarity with CLIP, retaining samples of high correspondence with a semantic score exceeding a threshold of 0.25. For samples with similarity scores below this specified threshold, we apply data augmentation techniques such as paraphrasing to revise captions, thereby improving alignment between images and texts and enhancing the overall utilization of the data. 
% The data processing flow is illustrated in Figure \ref{fig:data_clean_pipeline}.

\textbf{Data Augmentation.} Building upon the findings of Nguyen et al. (2023)\cite{nguyen2023improving} that synthetic captions can be used to improve multi-modal dataset quality, we propose a method to generate captions with improved relevance to the associated images in our BM-6B dataset.   To achieve this, we aim to enhance the contextual alignment and coherence between raw and synthetic captions by training a rewritten model. As a first step, we utilize the BLIP2\cite{blip2} model to generate synthetic captions for each image. Then, annotators are provided with guidelines to revise the original captions by considering the image content and the BLIP2-generated captions, and they are encouraged to preserve terms from the raw captions where appropriate, ensuring that these terms do not introduce misleading or incorrect details that are not evident in the image.   Subsequently,  we adapt the BLIP2 model into a rewritten model that is fine-tuned to take the image and original captions as input, and to produce the refined annotations as output. Once the fine-tuning process is complete, the resulting model is ready to generate improved captions for image-text pairs with low CLIP scores.

Table ~\ref{tab:training_data_distribution} displays the quantities of data remaining from each source dataset after the application of data cleaning and data augmentation procedures.

% \begin{figure*}
%     \centering
%     \includegraphics[width=1\linewidth]{data_flow_v2.png}
%     \caption{Enter Caption}
%     \label{fig:data_clean_pipeline}
% \end{figure*}

\begin{figure*}[tbh]
    \centering
    \includegraphics[width=1\linewidth, height=0.5\linewidth]{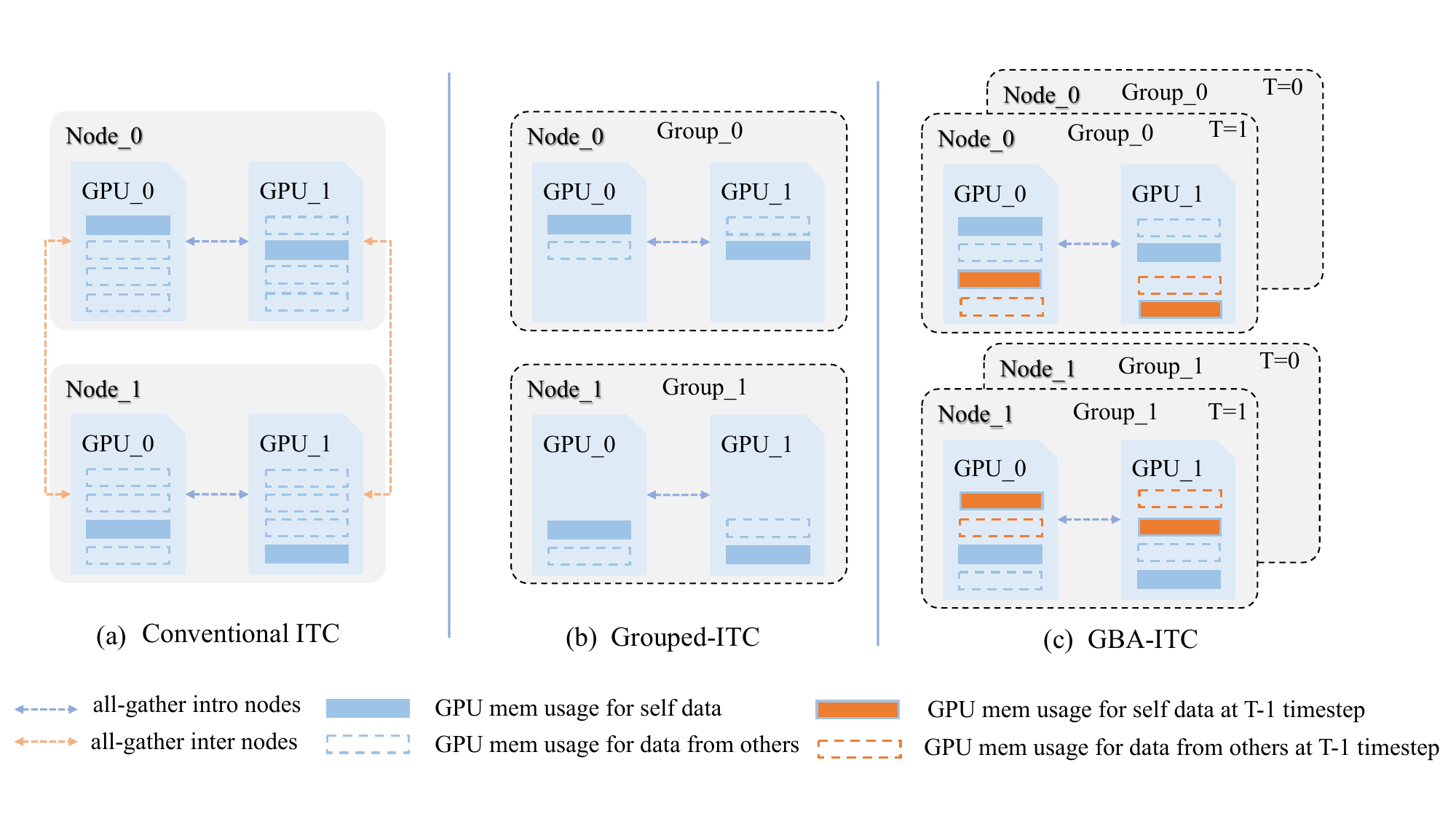}
    \caption{Illustration of conventional ITC, Grouped-ITC and GBA-ITC's data collection process before loss calculation with 2 nodes. GBA-ITC demonstrates the least communication overload.}
    \label{fig:GBA_ITC}
\end{figure*}

\subsection{Training Infrastructure}

Our training cluster consists of up to 32 NVIDIA DGX nodes, each equipped with 8 Ampere A100-SXM3-80G GPUs. In general, the application of training strategies such as PyTorch's Fully Sharded Data Parallel (FSDP)\footnote{https://pytorch.org/docs/1.11/fsdp.html} for parameter sharding is feasible for large-scale model training, since it demonstrates horizontal scalability of training resources.  However, our model involves the computation of  ITC loss.  Conventional computation of ITC loss, as illustrated in Figure \ref{fig:GBA_ITC} (a), necessitates the collection of image-text representations from all GPUs within the distributed system at each training step. This requirement engenders two significant challenges: 1. Frequent invocation of the all-gather operation across all nodes during training can result in a communication bandwidth bottleneck, thereby impeding efficient scaling in large-scale training scenarios. 2. As training scales, there is a linear increase in the volume of data aggregated on each GPU, leading to elevated peak memory usage and a higher risk of GPU memory depletion, especially with larger overall batch sizes.

We propose a new grouped aggregation strategy dubbed Grouped-ITC with batch accumulation (abbreviated as GBA-ITC) to address these issues.   The idea of Grouped-ITC is inspired by FSDP so that one can have multiple FSDP units, with the all-gather operation being executed only within each FSDP unit, thus minimizing unnecessary data movement and leading to reduced communication overhead. Our proposed Grouped-ITC inherits this concept by evenly partitioning the cluster nodes into multiple groups and then performing all-gather within each group during the ITC loss calculation. We illustrate this process with a 2 nodes example shown in Figure \ref{fig:GBA_ITC}(b), after dividing all nodes into two groups, by implementing Grouped-ITC, the peak GPU memory requirement (2 units per GPU) is reduced by half compared to conventional ITC (4 units per GPU), as well as reducing the all-gather communication overhead. Grouped-ITC's reduction in peak GPU memory usage enables accumulating training samples through multiple forward passes with a single gradient backpropagation step. By combining Grouped-ITC with batch accumulation, we achieve increased batch sizes and enhanced training efficiency. This method is designated as GBA-ITC and is depicted in Figure \ref{fig:GBA_ITC}(c).

\section{Experiments}

We train $M^2$-Encoder models of various sizes and evaluate them as follows. Detailed model configurations and training settings are provided in Appendix \ref{app:implementation_details}.

\subsection{Evaluation Setting}

\textbf{Coarse-grained Understanding.} To evaluate our image-text coarse understanding capabilities, we utilize several open Chinese and English retrieval and classification datasets. For English understanding, we assess zero-shot image-text retrieval capabilities on the Flickr30K \cite{plummer2015flickr30k} and COCO \cite{chen2015microsoft} datasets, and zero-shot image classification on ImageNet \cite{deng2009imagenet}, using the experimental setup described in CoCa \cite{coca}. For Chinese understanding, in line with evaluation method used by CN-CLIP \cite{cnclip}, we employ Flickr30K-CN and COCO-CN to test zero-shot image-text retrieval, and ImageNet-CN to evaluate zero-shot image classification capabilities.

\textbf{Fine-grained Understanding.} To evaluate fine-grained capabilities-including recognition of fine-grained categories, counting abilities, recognition of multiple object combinations, and understanding of relationships between objects-we constructed a fine-grained benchmark comprising seven diverse datasets. For a detailed description of the datasets included in this benchmark, readers are referred to Appendix~\ref{app:fine_grained_details}.

\textbf{Evaluation Metrics.} For tasks such as image-text retrieval and fine-grained retrieval, unless otherwise stated, we use the Mean Recall at top-1, top-5, and top-10 ranks for both text-to-image and image-to-text retrieval as our primary evaluation metrics. For the image classification task, we report the top-1 accuracy, which refers to the top-1 recall in a zero-shot setting.

\begin{table}[bpth]
	\centering
	\resizebox{1.0\linewidth}{!}{
		\begin{tabular}{l|c}
			\toprule
			Method & Accuracy  \\
			\midrule
			CLIP-0.4B \cite{radford2021learning} & 76.2\\
			ALIGN-0.8B \cite{jia2021scaling} & 76.4  \\
			FILIP-0.4B \cite{yao2021filip} & 78.3 \\
                Florence-0.9B \cite{yuan2021florence}& 83.7 \\
                LiT-1B \cite{zhai2022lit} & 84.5 \\
                BASIC-0.4B \cite{pham2021combined} & 81.5 \\
                BASIC-3B \cite{pham2021combined} & 85.7 \\
                OpenCLIP-1.3B\cite{cherti2023reproducible} & 78.5 \\
                OpenCLIP-2.5B\cite{cherti2023reproducible} & 80.0 \\
                CoCa-0.4B \cite{coca}& 82.6  \\
                CoCa-0.8B \cite{coca} & 84.8 \\
                CoCa-2.1B \cite{coca} & \underline{86.3} \\
                EVA-1.1B \cite{sun2023eva} & 78.5 \\
                EVA-5.0B \cite{sun2023eva}& 82.0 \\
                \hline
                $M^2$-Encoder-0.4B & 78.5 \\
                $M^2$-Encoder-1B & 85.4 \\
                $M^2$-Encoder-10B & \textbf{88.5} \\
			\bottomrule
		\end{tabular}
	}
	\caption{Zero-shot image classification evaluation results on ImageNet. Metrics denoted by an underscore represent the previous SOTA, while metrics presented in bold indicate surpassing previous SOTA.}
	\label{english zero-shot classification}
\end{table}

\begin{table}[bpth]
	\centering
	\resizebox{1.0\linewidth}{!}{
		\begin{tabular}{l|c}
			\toprule
			Method & Accuracy\\
			\midrule
BriVL-1B \cite{huo2021wenlan} &  24.3 \\
Taisu-0.2B \cite{liu2022taisu} & 34.0 \\
Wukong \cite{gu2022wukong} & 56.9 \\
CN-CLIP-0.2B \cite{cnclip} & 48.3 \\
CN-CLIP-1B \cite{cnclip} & \underline{59.6} \\
\hline
$M^2$-Encoder-0.4B & \textbf{69.1} \\
$M^2$-Encoder-1B & \textbf{78.5} \\
$M^2$-Encoder-10B & \textbf{80.7} \\
			\bottomrule
		\end{tabular}
	}
	\caption{Zero-shot image classification evaluation results on ImageNet-CN. Metrics denoted by an underscore represent the previous SOTA, while metrics presented in bold indicate surpassing previous SOTA.}
	\label{chinese zero-shot classification}
\end{table}

\begin{table*}[bpth]
	\centering
	\resizebox{1.0\linewidth}{!}{
		\begin{tabular}{l|cccccccc}
			% \hline
			% & \multicolumn{7}{c|}{CN} \\
			\cmidrule(r){1-9}
			%Method & \multicolumn{3}{c}{Text $\rightarrow$ Image} & \multicolumn{3}{c}{Image $\rightarrow$ Text} & \multicolumn{1}{c|}{~} & \multicolumn{3}{c}{Text $\rightarrow$ Image} & \multicolumn{3}{c}{Image $\rightarrow$ Text} & \multicolumn{1}{c}{~}\\
			% \cmidrule(r){2-4} \cmidrule(r){5-8} \cmidrule(r){9-11} \cmidrule(r){12-15} 
			Method(CN/EN) & CUB-200-2011 & Stanford-Dogs & CARS196 & COCO-MCC & COCO-COUNT & VG-POS &HOI-POS & Overall MR\\
            \hline
                \tabincell{c}{CLIP-0.4B \\ \cite{radford2021learning}} & -/81.56 &-/75.71 & -/87.98 & -/22.42 & -/17.87 & -/13.33 & -/63.31 & -/\underline{51.74}\\
                \tabincell{c}{CN-CLIP-1B \\ \cite{cnclip}} &24.16/- & 48.81/- & 83.58/- & 20.6/- & 14.54/-  & 9.96/- & 68.51/- & \underline{38.59}/-\\
			\midrule

   $M^2$-Encoder-0.4B & 73.72/57.22&85.19/84.66&94.14/91.54 & 41.27/38.54&25.48/24.47&23.74/21.04 & 75.44/70.94 & \textbf{59.85}/\textbf{55.49}\\
   $M^2$-Encoder-1B & 79.48/68.48&91.04/91.95&94.14/94.21 & 48.24/46.71&29.18/31.27&27.39/25.77& 81.21/79.10  &\textbf{64.38}/\textbf{62.50}\\
   $M^2$-Encoder-10B & 89.56/86.62&92.5/91.97&97.41/97.80&55.13/50.91&36.04/33.77 &32.78/26.96 &  84.45/80.57 & \textbf{69.70}/\textbf{66.94}\\
   \hline
		\end{tabular}
	}
	\caption{Zero-shot fine-grained retrieval evaluation results on 7 Chinese and English datasets. Metrics denoted by an underscore represent the previous SOTA, while metrics presented in bold indicate surpassing previous SOTA.}
	\label{fine-grained zero-shot retrieval}
 
\end{table*}

\subsection{Main Results}

\textbf{Image Classification.} The zero-shot image classification results for English and Chinese benchmarks are shown in Table~\ref{english zero-shot classification} and Table~\ref{chinese zero-shot classification}, respectively. In the English benchmark, our $M^2$-Encoder-0.4B and $M^2$-Encoder-1B outperform other methods with a similar number of parameters, while the $M^2$-Encoder-10B achieves SOTA results on ImageNet. For Chinese image classification, as detailed in Table~\ref{chinese zero-shot classification}, our $M^2$-Encoders surpass all existing methods. These experimental results underscore our method's effectiveness in both English and Chinese image classification tasks.

\textbf{Image-Text Retrieval.} The zero-shot retrieval results on English and Chinese benchmarks are shown in Table~\ref{english zero-shot retrieval} and Table~\ref{chinese zero-shot retrieval}, respectively. From Table ~\ref{english zero-shot retrieval}, we can observe that our $M^2$-Encoders exhibit superior performance in English image-text retrieval tasks. Specifically, $M^2$-Encoder-0.4B surpasses models with an equivalent number of parameters. $M^2$-Encoder-1B outperforms all existing methods, including those with larger parameter sizes such as OpenCLIP-2.5B, CoCa-2.1B and EVA-5.0B. Additionally, our $M^2$-Encoder-10B achieves the best results, with a gain in MR of 2.0\% on Ficker30K and 1.2\% on COCO. For Chinese image-text retrieval, the results are presented in Table~\ref{chinese zero-shot retrieval}. Our $M^2$-Encoders outperform all other methods across a range of model parameters on both Flickr30K-CN and COCO-CN.

\begin{table*}[t]
	\centering
	\resizebox{0.95\linewidth}{!}{
		\begin{tabular}{l|ccccccc|ccccccc}
			\hline
			& \multicolumn{7}{c|}{Flickr30K} & \multicolumn{7}{c}{MSCOCO} \\
			\cmidrule(r){2-8} \cmidrule(r){9-15}
			Method & \multicolumn{3}{c}{Text $\rightarrow$ Image} & \multicolumn{3}{c}{Image $\rightarrow$ Text} & \multicolumn{1}{c|}{~} & \multicolumn{3}{c}{Text $\rightarrow$ Image} & \multicolumn{3}{c}{Image $\rightarrow$ Text} & \multicolumn{1}{c}{~}\\
			\cmidrule(r){2-4} \cmidrule(r){5-8} \cmidrule(r){9-11} \cmidrule(r){12-15} 
			& R@1 & R@5 & R@10 & R@1 & R@5 & R@10 &MR& R@1 & R@5 & R@10 & R@1 & R@5 & R@10 & MR\\
			\midrule
   CLIP-0.4B \cite{radford2021learning}& 68.7&90.6&95.2 &88.0&98.7& 99.4 &90.1 & 37.8 & 62.4 &72.2 &58.4&81.5&88.1 &66.7\\
   ALIGN-0.8B \cite{jia2021scaling} & 75.7&93.8&96.8 & 88.6&98.7&99.7 &92.2 & 45.6&69.8&78.6 & 58.6&83.0&89.7 &70.9\\
   FLAVA-0.3B \cite{singh2022flava}& 65.2&89.4&- & 67.7&94.0&- &-& 38.4&67.5&- & 42.7&76.8&- &-\\
   FILIP-0.4B \cite{yao2021filip}& 75.0&93.4&96.3 & 89.8&99.2&99.8 &92.3& 45.9&70.6&79.3 & 61.3&84.3&90.4 &72.0\\
   Florence-0.9B \cite{yuan2021florence} & 76.7&93.6&-	- & 90.9&99.1&-&-&47.2&71.4&-& 64.7&85.9&- &-\\
   BEIT3-0.7B	\cite{wang2022image} & 81.5&95.6&97.8 & 94.9&99.9&100  &\underline{94.9} &- & -&-&-&-&- &-\\
   OpenCLIP-1.3B \cite{cherti2023reproducible} & 77.7&94.1&96.9 & 91.4&99.2&99.6 &93.1& 48.8&73.3&81.5 & 66.4&86.0&91.8 &74.6\\
   OpenCLIP-2.5B \cite{cherti2023reproducible}& 79.5&95.0&97.1 & 92.9&99.3&99.8 &93.9& 51.4&74.9&83.0 & 67.3&86.9&92.6 &76.0\\
   CoCa-0.4B \cite{coca}& 76.8&93.7&96.8 & 89.8&98.8&99.8 &92.6& 47.5&72.4&80.9 & 63.8&84.7&90.7 &73.3\\
   CoCa-0.8B \cite{coca} & 79.0&95.1&97.4 & 91.4&99.2&99.9 &93.7& 50.1&73.8&81.8 & 65.4&85.6&91.4 &74.7\\
   CoCa-2.1B \cite{coca} & 80.4&95.7&97.7	& 92.5&99.5&99.9 &94.3& 51.2&74.2&82.0 & 66.3&86.2&91.8 &75.3\\
   EVA-1.1B \cite{sun2023eva}& 72.6&91.6&95.1 & 88.3&98.3&99.3 & 90.9&44.1&68.5&77.3 & 61.8&83.3&90.0 &70.8\\
   EVA-5.0B \cite{sun2023eva}& 78.8&94.2&96.8 & 93.9&99.4&99.8 &93.8& 51.1&75.0&82.7 & 68.8&87.8&92.8 & 76.4\\
   ERNIE-ViL 2.0 \cite{shan2022ernie}& 77.4&93.8&96.4 & 91.2&99.1&99.8 &92.9& 46.0&71.4&80.4 & 63.1&85.7&91.4 &73.0\\
   AltCLIP-0.8B \cite{chen2022altclip} & 72.5&91.6&95.4 & 86.0 &98.0 &99.1 &  90.4 & 42.9 & 68.0 & 77.4 & 58.6 & 80.6 & 87.8 & 69.2 \\
   UMG-CLIP-0.4B \cite{shi2024umgclip} & 83.1 & 96.0& 98.1 & 93.4 &99.5 &99.9 & 95.0 & 54.6 & 78.5 & 86.1 & 68.9 &89.0 & 94.1 & 78.5\\
   UMG-CLIP-5B \cite{shi2024umgclip} & 81.2 & 95.7 & 97.7& 93.1 & 99.7 & 99.9&94.6&57.5 &80.4& 87.3&71.7 &89.8 &94.3& \underline{80.2}\\
   \hline
   $M^2$-Encoder-0.4B & 87.2&97.9&99.4 & 85.2&98.2&99.4 &94.5& 49.0&76.1&85.0 &62.5&86.4&92.7 &75.2\\
   $M^2$-Encoder-1B & 88.8&98.9&99.5 & 87.6&98.3&99.4 &\textbf{95.4} & 51.5&76.8&85.0 & 67.4&87.7&93.2 &76.9\\
   $M^2$-Encoder-10B & 92.2&99.5&99.7 & 91.2&99.2&99.6 &\textbf{96.9}&  56.5&81.6&88.8 & 72.8&92.3&96.3 & \textbf{81.4}\\
			\hline
		\end{tabular}
	}
	\caption{Zero-shot image-text retrieval evaluation results on Flickr30K and MSCOCO dataset. Metrics denoted by an underscore represent the previous SOTA, while metrics presented in bold indicate surpassing previous SOTA.}
	\label{english zero-shot retrieval}
\end{table*}

\begin{table*}[t]
	\centering
	\resizebox{0.95\linewidth}{!}{
		\begin{tabular}{l|ccccccc|ccccccc}
			\hline
			& \multicolumn{7}{c|}{Flickr30K-CN} & \multicolumn{7}{c}{COCO-CN} \\
			\cmidrule(r){2-8} \cmidrule(r){9-15}
			Method & \multicolumn{3}{c}{Text $\rightarrow$ Image} & \multicolumn{3}{c}{Image $\rightarrow$ Text} & \multicolumn{1}{c|}{~} & \multicolumn{3}{c}{Text $\rightarrow$ Image} & \multicolumn{3}{c}{Image $\rightarrow$ Text} & \multicolumn{1}{c}{~}\\
			\cmidrule(r){2-4} \cmidrule(r){5-8} \cmidrule(r){9-11} \cmidrule(r){12-15} 
			& R@1 & R@5 & R@10 & R@1 & R@5 & R@10 &MR& R@1 & R@5 & R@10 & R@1 & R@5 & R@10 & MR\\
			\midrule
   BriVL-1B \cite{huo2021wenlan} & 10.3&27.5&37.9 & 17.7&42.3&54.3 & 31.7 &  14.8&39.0&54.2 & 17.1&41.7&57.5 & 37.4\\
   Taiyi-0.3B \cite{luo2023taiyi} & 53.7&79.8&86.6 & 63.8&90.5&95.9 & 78.4 & - & -& - & - & -& -&-\\
   Wukong-0.4B \cite{gu2022wukong} &  51.7&78.9&86.3 & 76.1&94.8&97.5 & 80.9 & 53.4&80.2&90.1 & 55.2&81.0&90.6  & 75.1\\
   Taisu-0.2B \cite{liu2022taisu} & 49.9&78.9&87.0 & 65.6&90.1&94.9 & 77.7 & 53.6&83.7&92.4 & 52.5&81.5&91.4 & 75.9\\
   CN-CLIP-0.2B \cite{cnclip} & 62.7&86.9&92.8 & 74.6&93.5&97.1 & 84.6 & 57.0&84.1&93.6 & 62.2&86.6&94.9 & 79.7\\
   CN-CLIP-1B \cite{cnclip} & 71.2&91.4&95.5 & 81.6&97.5&98.8 & \underline{89.3} & 69.2&89.9&96.1 & 63.0&86.6&92.9 & 83.0\\
   R2D2-0.4B \cite{xie2022zero} & 60.9&86.8&92.7 & 77.6&96.7&98.9 & 85.6 & 56.4&85.0&93.1 & 63.3&89.3&95.7 & 80.5\\
   AltCLIP-0.8B \cite{chen2022altclip} & 69.8&89.9&94.7 & 84.8&97.4&98.8 & 89.2 &61.3 & 86.0& 93.2 & 77.8 & 94.4& 97.5&85.0\\   
   ERNIE-ViL 2.0 \cite{shan2022ernie} & -&- &-&-&-&-&-&69.6&91.2&96.9 &69.1&92.9&97.1 & \underline{86.1} \\
   \hline
   $M^2$-Encoder-0.4B & 72.5&92.4&96.4 & 87.9&98.4&99.4 & \textbf{91.2} & 72.5&93.7&97.7 &  71.0&93.7&97.9 & \textbf{87.8}\\
   $M^2$-Encoder-1B & 78.6&94.8&97.6 & 91.5&99.3&99.7 & \textbf{93.6} &75.9&95.1&98.3 & 74.6&94.2&98.0 & \textbf{89.3} \\
   $M^2$-Encoder-10B& 81.5&96.2&98.5 & 93.8&99.7&99.9 & \textbf{94.9} & 78.7&96.0&98.7 & 80.9&96.5&99.1 & \textbf{91.7}\\
			
   %Ours & \textbf{46.6} & \textbf{76.3} & \textbf{83.6} & \textbf{35.9} & \textbf{64.6} & \textbf{75.6} & \textbf{18.7} & \textbf{41.0} & \textbf{53.6} & \textbf{17.2} & \textbf{39.7} & \textbf{51.6} \\
			\hline
		\end{tabular}
	}
	\caption{Zero-shot image-text retrieval evaluation results on Flickr30K-CN and COCO-CN dataset. Metrics denoted by an underscore represent the previous SOTA, while metrics presented in bold indicate surpassing previous SOTA.}
	\label{chinese zero-shot retrieval}
\end{table*}

\textbf{Fine-grained Retrieval.} Our $M^2$-Encoders introduce token-level local alignment to enhance fine-grained capabilities. Table~\ref{fine-grained zero-shot retrieval} presents the fine-grained retrieval results on seven datasets, encompassing both Chinese and English languages. To highlight the fine-grained advantages of the $M^2$-Encoders, we used the open-source models CN-CLIP-1B \cite{cnclip} and CLIP \cite{radford2021learning} as baselines for Chinese and English retrieval, respectively. Our method exhibits a superior fine-grained retrieval performance, with even the smallest variant, $M^2$-Encoder-0.4B, outstripping the baseline models on the majority of the fine-grained datasets. This affirms the efficacy of our method in fine-grained understanding and modeling.

\subsection{Ablation Study}

\begin{table}[bth]
\centering
\resizebox{1.0\linewidth}{!}{
\begin{tabular}{c|c|c|c}
\hline
Setting & \tabincell{c}{Batch-size\\per GPU} & \tabincell{c}{Group-size\\(\#GPUs)}& \tabincell{c}{Batch-Acc\\steps}\\
\hline
ITC  & 512 & 32 & 1 \\
Grouped-ITC & 1024 & 16 & 1 \\
GBA-ITC & 1024 & 8 & 2 \\
\hline
\end{tabular}
}
\caption{Experimental configuration for conventional ITC, Grouped-ITC and GBA-ITC.  All ITC losses calculcation involve an equal number of samples. Group size means the number of GPUs involved in data aggregation for loss computation, and Batch-Acc steps represents for batch accumulation steps for each GPU.}
\label{tab:grouped_itc_exp}
\end{table}

\begin{figure*}
    \centering
    \includegraphics[width=0.8\linewidth, height=0.3\linewidth]{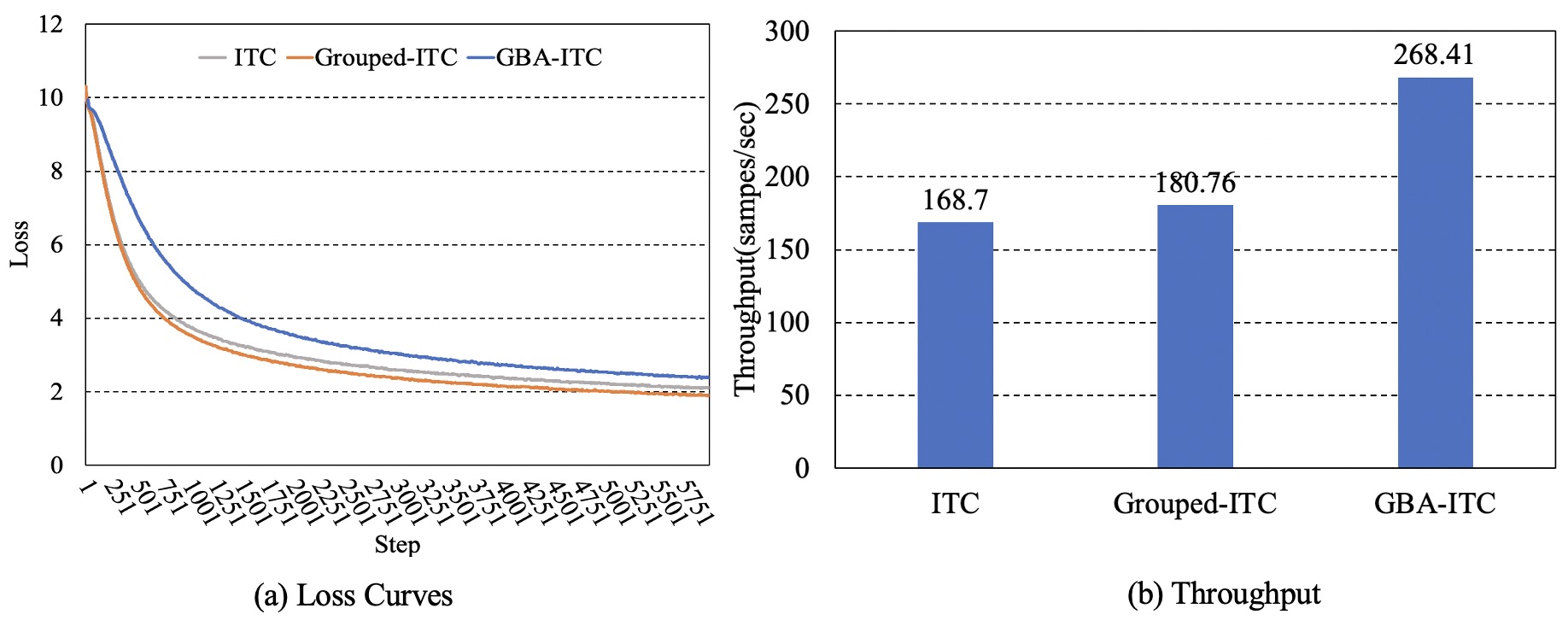}
    \setlength{\abovecaptionskip}{0.1cm}
    \caption{Comparison of loss curves and throughput among ITC, Grouped-ITC and GBA-ITC.}
    \label{fig:itc_comparison}
\end{figure*}

\begin{table}[t]
	\centering
	\resizebox{1.0\linewidth}{!}{
		\begin{tabular}{l|ccc|ccc}
			\hline
		\multirow{2}{*}{Method}	& \multicolumn{3}{c|}{Flickr30K-CN} & \multicolumn{3}{c}{COCO-CN}\\
			\cmidrule(r){2-4} \cmidrule(r){5-7} 
			 & T2I-MR & I2T-MR & MR &  T2I-MR & I2T-MR & MR \\
			\midrule
   $M^2$-Encoder-1B(33\%) & 85.6&94.6& 90.1 & 83.0 & 85.5 & 84.3 \\
   $M^2$-Encoder-1B(67\%) & 89.0&96.6& 92.8 & 88.9 & 89.6 & 89.2\\
   $M^2$-Encoder-1B(100\%)& 90.3&96.8& 93.6 & 89.8 & 88.9 & 89.3\\		
		\hline
		\end{tabular}
	}
	\caption{Evaluation results for pertaining $M^2$-Encoder-1B with different percentage of the proposed BM-6B dataset.}
	\label{tab:datascale_ablation}
\end{table} 

\begin{table*}[h]
	\centering
	\resizebox{1.0\linewidth}{!}{
		\begin{tabular}{l|cccccccc}
			\cmidrule(r){1-9}
			% \cmidrule(r){2-4} \cmidrule(r){5-8} \cmidrule(r){9-11} \cmidrule(r){12-15} 
			Method(CN) & CUB-200-2011 & Stanford-Dogs & CARS196 & COCO-MCC & COCO-COUNT & VG-POS &HOI-POS & Overall MR\\
            \hline
   $M^2$-Encoder-1B w/ ITC-only  & 68.52 &86.93 &90.93 & 40.44 &25.26 &22.62 & 76.28  &\textbf{58.71} \\
   $M^2$-Encoder-1B  & 79.48 & 91.04 & 94.14 & 48.24 &29.18& 27.39 & 81.21 & \textbf{64.38}\\
   \hline
		\end{tabular}
	}
	\caption{Ablation on the effectiveness of pretraining tasks for fine-grained understanding.}
	\label{tab:finegrained_ablation}
\end{table*}

\textbf{Benefits of Large Training Sets}. We studied the impact of the size of our BM-6B pretraining dataset.
For this experiment, we take subsets of approximately 33\% and 67\% from our BM-6B dataset and pre-train the $M^2$-Encoder-1B model using the same setup as above. As shown in Table\ref{tab:datascale_ablation}, we can observe that the performance of $M^2$-Encoder-1B grows monotonically as the usage of image-text pairs from BM-6B increases, which suggests that our $M^2$-Encoders may benefit from even more pretraining data.

\textbf{Effectiveness of Pretraining Tasks}. The enhanced fine-grained understanding ability makes our $M^2$-Encoder series models distinct from existing ones. To validate the effectiveness of our pretraining tasks in this domain, we trained an $M^2$-Encoder-1B variant(dubbed ITC-only) employing only ITC loss employed as a baseline for comparative analysis. The results are demonstrated in Table \ref{tab:finegrained_ablation}. With the aid of our fine-grained pertaining loss, our $M^2$-Encoder-1B outperforms its ITC-only counterpart by a large margin on tasks requiring detailed perception, illustrating the effectiveness of our pretraining approach on fine-grained understanding.

\textbf{Effectiveness of GBA-ITC}. To verify the effectiveness of GBA-ITC, comparative experiments (detailed in Table \ref{tab:grouped_itc_exp}) were conducted on both performance and effectiveness using 4 DGX nodes (8GPUs per node). Results are shown in Figure \ref{fig:itc_comparison}. It can be observed that under the same global batch size setting,  both Grouped-ITC and GBA-ITC exhibit convergence behavior analogous to that of the conventional ITC with regard to loss metrics. The observed minor fluctuations in loss metrics may be attributed to batch accumulation operations in GBA-ITC, which result in less frequent optimization and parameter updates over an equivalent number of iterations. Regarding throughput, the Grouped-ITC method, which performs aggregation within each group, incurs lower communication overhead compared to the conventional ITC baseline. This efficiency translates to a throughput improvement factor of 1.07X. Additionally, GBA-ITC leverages smaller group sizes coupled with batch accumulation to effectively diminish communication costs, culminating in a further throughput enhancement of 1.59X. The group size of GBA-ITC can be flexibly configured, enabling the decoupling of the ITC loss computation from the overall batch size. In our experiment with same batch size per GPU, reducing the group size from 32 to 16 resulted in peak memory usage being reduced by approximately half( from 50.42GB to 27.46GB). These findings indicate that GBA-ITC is effective in mitigating the challenges of communication bottlenecks and high peak memory usage, thus facilitating efficient scaling in large-scale training scenarios.

\section{Related Work}
% Contrastive-based vision-language models (VLMs) serve key roles in multimodal Large Language Models\cite{mplugowl,blip2,gong2023multimodal,minigpt4,instructblip,llava,qwen-vl} and cross-modal generation models\cite{dalle2,dalle1,sd}. 
Recent advancements in adapting VLMs for Chinese language understanding include CN-CLIP\cite{cnclip} and AltCLIP\cite{chen2022altclip}. CN-CLIP enhances CLIP\cite{clip} with Chinese language support by utilizing locked-image tuning\cite{lit} to keep the CLIP visual Encoder constant while aligning it with a Chinese text encoder in the first stage, followed by contrastive fine-tuning using a dataset of 200 million Chinese image-text pairs in the second stage. Meanwhile, AltCLIP extends CLIP with Chinese support by aligning the CLIP text encoder with a multilingual text encoder using a teacher-learning approach. 
% Both CN-CLIP and AltCLIP have achieved significant improvements in the comprehension abilities of VLMs for the Chinese language. However, due to limitations in their model sizes and the lack of large-scale Chinese image-text datasets, there is still substantial room for improvement in Chinese VL understanding.
Our approach differs from the methods mentioned above in three key aspects: Firstly, unlike CN-CLIP and AltCLIP, which build upon the existing CLIP model, our bilingual $M^2$-Encoders are developed without relying on any pre-existing pretrained models, directly trained from scratch using the massive bilingual BM-6B dataset. Secondly, these CLIP-based models tend to underperform on tasks that require detailed perception, since they rely on using only the ITC task for cross-modal alignment. Our $M^2$-Encoders are trained with enhanced fine-grained understanding capability. Thirdly, CN-CLIP and AltCLIP are limited to a maximum model size of 1B parameters, potentially constraining their ability to capture intricate patterns. Our scalable model architecture and the BM-6B dataset have enabled us to train a model with up to 10 billion parameters. This has resulted in setting new state-of-the-art benchmarks in both Chinese and English multimodal tasks and is known to be the largest-scale bilingual contrastive-based vision-language model to date.
% Secondly, since our $M^2$ Encoders are trained from scratch, we can use current cutting-edge architectures such as the MAGNETO\cite{magneto} transformer, thus facilitating model scaling in contrast to conventional CLIP architecture. 

\section{Conclusion}
In this work, we propose $M^2$-Encoders, a series of bilingual vision-language foundation models with multiple sizes ranging from 0.4B to 10B parameters that facilitate both coarse and fine-grained understanding. To adequately train the $M^2$-Encoders from scatch, we construct the BM-6B, an ultra-large bilingual pretraining dataset comprising 6 billion image-text pairs with Chinese and English data nearly equally distributed, addressing the need for diverse and extensive bilingual datasets. To facilitate efficient scaling in large-scale training, we introduce the GBA-ITC method for reduced communication overhead and high peak memory usage. Our comprehensive evaluation shows that $M^2$-Encoders,  particularly the 10B variant, set new benchmarks in both languages for multimodal retrieval and classification tasks. Furthermore, we underscore that $M^2$-Encoders can also achieve competitive performance in zero-shot fine-grained retrieval across 7 diverse datasets.

\bibliography{anthology,custom}
\bibliographystyle{acl_natbib}

% \clearpage
\appendix
\section{Implementation Details}
\label{app:implementation_details}
Configurations of our $M^2$-Encoders of different sizes are shown in Table ~\ref{tab:model_configurations}. We train $M^2$-Encoder series with proxy tasks introduced in Section \ref{sec: model}, all models are trained from scratch on BM-6B and we adopt slightly different training settings optimized across model scales as shown in Table ~\ref{tab:training_setting}.

\begin{table}[btph]
	\centering
	\resizebox{1.0\linewidth}{!}{
		\begin{tabular}{c|ccc|c}
			\toprule
			 \multirow{2}{*}{Models} & \multicolumn{3}{c|}{Encoder} &  \multirow{2}{*}{\# Params}  \\
                \cmidrule(r){2-4}
			 & Layers & Width & Heads\\
			\midrule
                $M^2$-Encoder-0.4B & 12 & 768 & 12 & 402M\\
                $M^2$-Encoder-1B & 24 & 1024 & 16 & 924M\\
                $M^2$-Encoder-10B & 24 & 4096 & 32 & 11.2B\\
			\bottomrule
		\end{tabular}
	}
	\caption{Architecture configurations.}
	\label{tab:model_configurations}
\end{table}

% TODO: 训练轮数
\begin{table}[btph]
	\centering
	\resizebox{1.0\linewidth}{!}{
		\begin{tabular}{c|c}
			\toprule
			Hyperparameter & 0.4B or 1B / 10B\\
			\midrule
   Image-text data & BM-6B \\
   Proxy loss balance & $\alpha$=$\beta$=0.3\\
   Learning rate & 2e-4 / 4e-4 \\
   Optimizer & LAMB \cite{you2019large} \\
   Optimizer parameters & $\beta_1$, $\beta_2$, $\epsilon$ = 0.9, 0.98, 1e-6 \\
   Weight decay & 0.05 \\
   $\alpha$, $\beta$ & 0.3 \\
   Input resolution & $224^2$ \\
   Patch size & $16^2$ \\
   Overall Batch size  & 131072 / 143360 \\
   GBA-ITC Batch size & 32768 \\
   Samples seen & 9B \\
   Numerical Precision & DeepSpeed bf16 \\
   ZeRO Optimization & \tabincell{c}{Partial Redundancy Optimizer\\ (PaRO)\cite{wu2023rethinking}} \\
			\bottomrule
		\end{tabular}
	}
	\caption{Training settings for $M^2$-Encoders of various sizes.}
	\label{tab:training_setting}
\end{table}

\section{Fine-grained Benchmark Details}
\label{app:fine_grained_details}
The English portion of our fine-grained benchmark is derived directly from an open-source dataset, while the Chinese part is created through translation. The fine-grained benchmarks are listed in Table ~\ref{tab:fine-grained benchmark}. For CUB-200-2011 \cite{wah2011caltech}, Standford-Dogs \cite{khosla2011novel}, and CARS196 \cite{krause20133d} datasets, we use the images and labels from their respective test sets. VG-POS and HOI-POS are designed to represent orientation and interaction relationships, respectively, and consist of selected positive examples from the Visual Genome \cite{krishna2017visual} and HOI \cite{chao2015hico} datasets. Both COCO-MCC and COCO-Count are derived from the COCO \cite{chen2015microsoft} test set, where COCO-MCC is a multiclass classification dataset that includes fine-grained captions describing the classes of objects in each image, and COCO-Count is constructed specifically to assess the counting abilities of vision-language foundation models, with each image's caption detailing the types and quantities of objects present. The distribution  of our constructed fine-grained benchmark is shown in Table \ref{tab:fine-grained benchmark}.

\begin{table}[btph]
	\centering
	\resizebox{0.8\linewidth}{!}{
		\begin{tabular}{l| c| c}
			\toprule
			Dataset & \# Images & \# Texts\\
			\midrule
                CUB-200-2011 & 5794 & 200 \\
                Stanford-Dogs & 8580 & 120 \\
                CARS196 & 4021 & 196 \\
                COCO-MCC & 2222 & 2222 \\
                COCO-COUNT & 3240 & 3238\\
                VG-POS & 3000 & 53806\\
                HOI-POS & 520 & 8593\\
			\bottomrule
		\end{tabular}
	}
	\caption{Data distribution of our fine-grained benchmark.}
	\label{tab:fine-grained benchmark}
\end{table}

\section{Distributions of BM-6B data sources.}
\label{app:bm_6B_distributions}
Most of our collected bilingual image-text pairs are from publicly available sources, here we detail the data sources of our constructed BM-6B in Table \ref{tab:training_data_distribution}.

\begin{table}[bhpt]
    \centering
    \resizebox{0.95\linewidth}{!}{
    \begin{tabular}{c|c|c|c|c|c} \hline 
         Language&  Dataset&  Original& Sampled &Reamining & \#image-text pairs \\  \hline 
         \multirow{3}{*}{English}&  LAION-EN\cite{laion5b}&  2.3B&  1.9B&82.6\% & \multirow{3}{*}{3.07B}\\ 
         &  COYO-700M\cite{kakaobrain2022coyo-700m}&  700M&  435M&62.1\%&\\ 
 & DataComp-1B\cite{gadre2023datacomp}& 1.4B& 737M&52.6\%&\\ \hline 
 
 \multirow{7}{*}{Chinese}& LAION-CN\cite{laion5b}& 140M& 104M&74.3\%& \multirow{7}{*}{3.01B}\\ 
 & Wukong\cite{gu2022wukong}& 100M& 90M&90.0\%&\\ 
 & Taisu\cite{liu2022taisu}& 166M& 94M&56.6\%&\\  
 & Zero\cite{xie2022zero}& 250M& 152M&60.8\%&\\  
& LAION-EN translated&-&1.9B&-&\\ 
 & In-house Data& -& 668M&-&\\ \hline
    \end{tabular}
    }
    \caption{Distributions of BM-6B data sources.}
    \label{tab:training_data_distribution}
\end{table}

\end{document}